\documentclass[11pt]{article}

\usepackage[preprint]{acl}

\usepackage{times}
\usepackage{latexsym}
\usepackage[T1]{fontenc}
\usepackage[utf8]{inputenc}
\usepackage{microtype}
\usepackage{inconsolata}
\usepackage{graphicx}
\usepackage{booktabs}
\usepackage{multirow}
\usepackage{xcolor}
\usepackage{listings}
\usepackage{tikz}
\usetikzlibrary{tikzmark, fit, calc, arrows.meta}

\usepackage{cleveref}

\newcommand{\mknote}[1]{\textcolor{red}{#1}}
\renewcommand{\mknote}[1]{} %

\newcommand{\methodname}{\textsc{Lacuna}}

\newcommand{\btHL}[1]{\colorbox{yellow!20}{#1}}
\definecolor{errred}{RGB}{170,0,0}
\lstdefinelanguage{dotty}{
  basicstyle=\footnotesize\ttfamily,
  keywords={erased, val, var, if, then, in, handle,
    return, def, match, case, new, type, trait,
    package, object, given, eff,
    pretype, class, extends, extension, infix, else,
    box, unbox, try, catch, import, throw, throws, using, enum,
    use, cap, this, abstract, final,
    sealed, override, private, protected, transparent, inline, opaque, open, tracked, lazy, uses, uses_init, update, consume, while, do, summon, agent, agentSafe, eval, evalSafe, scala, with, true, false, null, for, yield, tailrec},
  keywordstyle=\bfseries\color{magenta!80!black},
  morekeywords=[2]{@assumeSafe, @untrackedCaptures, @use, @evalLike, @evalSafeLike, @tailrec},
  keywordstyle=[2]{\color{teal}},
  sensitive=true,
  comment=[l]{//},
  morecomment=[s]{/*}{*/},
  commentstyle=\color{green!40!black},
  stringstyle=\color{green!60!black},
  morestring=[b]',
  morestring=[b]",
  moredelim=**[is][\btHL]{`}{`},
  moredelim=[is][\color{errred}]{<!}{!>},
  columns=fullflexible,
  alsoletter={@},
}
\lstnewenvironment{code}
  {\lstset{language=dotty,
    basicstyle=\footnotesize\ttfamily,
    backgroundcolor=\color{gray!4},
    frame=trbL, framerule=0.0pt,
    rulecolor=\color{blue!20!gray!40},
    rulesep=1.5pt,
    rulesepcolor=\color{blue!20!gray!40},
    framesep=0pt,
    framexleftmargin=5pt, framexrightmargin=1pt,
    framextopmargin=3pt, framexbottommargin=3pt,
    xleftmargin=6pt,
    columns=flexible, keepspaces=true, showstringspaces=false,
    extendedchars=true, breaklines=false,
    escapeinside={(*}{*)},
    aboveskip=3pt, belowskip=1pt}}
  {}

\newcommand{\holeframe}[6]{%
  \par\nointerlineskip
  \vbox to 0pt{\vss
  \begin{tikzpicture}[remember picture, overlay]
    \draw[draw=#5, rounded corners, line width=0.7pt]
      let \p1=(pic cs:#1), \p2=(pic cs:#2), \p3=(pic cs:#3), \p4=(pic cs:#4) in
      ($(\x1,\y2) + (-3pt, -0.4ex)$) rectangle
      ($(\x3,\y4) + (4pt, -0.6ex)$);
    \path
      let \p2=(pic cs:#2), \p3=(pic cs:#3) in
      node[anchor=east, text=#5, fill=white, inner xsep=2pt, inner ysep=1pt,
           font=\scriptsize\itshape]
      at ($(\x3,\y2) + (4pt, -0.4ex)$) {#6};
  \end{tikzpicture}%
  }%
}

\title{\methodname{}: Safe Agents as Recursive Program Holes}

\author{
  \bfseries Yaoyu Zhao\thanks{Equal contribution.} \hspace{1.5em} Yichen Xu\footnotemark[1] \hspace{1.5em} Oliver Bračevac \\[4pt]
  \bfseries Cao Nguyen Pham \hspace{1.5em} Frank Zhengqing Wu \hspace{1.5em} Martin Odersky \\[10pt]
  \normalfont\normalsize EPFL, Lausanne, Switzerland \\[3pt]
  \normalfont\small\texttt{\{yaoyu.zhao, yichen.xu, oliver.bracevac,} \\
  \normalfont\small\texttt{nguyen.pham, zhengqing.wu, martin.odersky\}@epfl.ch}
}

\makeatletter%
\begin{document}
\maketitle
\begin{abstract}

	LLM agents increasingly act by writing code, yet a split
	persists between the runtime that drives the agent and the code
	the model writes. The runtime owns the loop, context, and control
	flow, and the model has little say over any of them.
	Letting model-written code shape the runtime itself would make
	agents more expressive, but it would also sharpen safety problems.
	A model can be diverted by a prompt injection, call the wrong tool,
	or fail partway and leave an inconsistent state, and each such
	failure reaches further when the code shapes the runtime than when
	it expresses a single action.
	We present \methodname{}, a programming model for agents that closes this
	split while preserving safety.
	Each agent action is a typed call \lstinline|agent[T](task)| that
	the LLM fills with code when execution reaches it, and the code is
	type-checked against the surrounding program before it runs.
	Because each action is accepted or rejected as a whole, a rejected
	one leaves the environment untouched, and its compiler diagnostics
	drive a retry.
	The same check also bounds which tools and data an action may use
	and how they flow.
	Our primitive expresses ReAct loops,
	sub-agents, skills, parallel decomposition, and multi-model
	planning as ordinary control flow.
	We evaluate \methodname{}
	on a collection of test cases, BrowseComp-Plus, and
	$\tau^2$-bench. On BrowseComp-Plus, \(8.6\%\) of generations
	are rejected before execution, with \(0.7\) retries per query
	on average, and the agent reaches \(27.1\%\) accuracy. On
	$\tau^2$-bench, \methodname{} solves \(76.0\%\) of \(392\)
	tasks across four domains with a capable model,
	on par with the baseline agent.

\end{abstract}

\section{Introduction}
\label{sec:intro}

Large language models (LLMs) increasingly drive \emph{agents}:
programs that call models, use tools, and maintain state
to solve tasks. Tools such as file access, web search, and API
calls are now often described through protocols such as
MCP~\cite{mcp} and packaged as reusable skills~\cite{agentskills}.
The dominant approach, ReAct~\cite{DBLP:conf/iclr/YaoZYDSN023}, has
the model alternate between reasoning and individual tool calls
until reaching an answer. Code-as-action
agents~\cite{codeact,codeexecution,smolagents} offer an alternative:
instead of emitting one tool call at a
time, the model writes code that composes tools, parses intermediate
results, branches, and loops.

Existing code-as-action agents keep a clear split between the code
that \emph{runs} the agent and the code it \emph{writes}.
The runtime owns the loop, context, and action dispatch, while
the model supplies only the next fragment, with little say over what
context to keep, when to spawn sub-agents, or how to adapt control
flow.
Recursive language modeling~\cite{rlm}
lets generated code update a persistent execution context and call
the model again, but the runtime still owns the loop and call
structure.
Letting the code the model writes shape the runtime itself
lifts these limits and makes agents more expressive, but it also
raises the stakes for safety.
Model-written code is untrusted. When it only expresses actions, the
surrounding runtime bounds its reach; once the agent shapes its own
runtime, an attack reaches the runtime itself.
Existing defenses are piecemeal.
Sandboxes and restricted interpreters~\cite{monty} limit what
code can do at runtime, policy languages~\cite{cedar} gate access to
resources, and input-hardening and mediation
defenses~\cite{DBLP:conf/uss/ChenPSW25,dualllm} try to block unsafe
actions.
None of them checks a whole generated action before it starts.

We propose \methodname{}, a programming model that closes this
split while preserving safety.
Each agent action is a typed hole that
the LLM fills with code when execution reaches it, and the code is
type-checked against the surrounding program before it runs.
The core idea is to put the model call inside the program at the
point where its result is needed, and to make the caller state what
kind of result is expected.
Our prototype uses Scala~3 because it supports the two foundations we
need: compiling a fresh snippet in the surrounding program context,
and tracking which resources that snippet is allowed to
use~\cite{dottycc}.
The user-facing call is:
\begin{code}
def agent[T](task: String): T
\end{code}
\noindent Here \lstinline|task| is the natural-language prompt and
\lstinline|T| is the expected result type. When execution reaches
this call, the model writes Scala code for the request, which
\methodname{} compiles at the same point in the surrounding
program, so it can use the variables, functions, and tools available
there. If the code provably produces a value of type \lstinline|T|,
it runs; otherwise the compiler's errors are sent back as feedback
for a retry.

The generated code is ordinary Scala, not just a single tool call.
It can use tools, process data, and ask the model for help again,
as in this request for a report over several topics:
\begin{code}
val topics = List(
  "LLM", "world models", "transformer", ...)
val report =(*\tikzmark{t1}*)
  (*\tikzmark{c1a}*)agent[String](
    "Research each topic and generate a" +(*\tikzmark{c1r}*)
    " report on their connections.")(*\tikzmark{c1b}*)
\end{code}
\holeframe{c1a}{t1}{c1r}{c1b}{blue!55!black}{typed action}

\noindent One valid expansion uses nested research calls:
\begin{code}
val report: String = (*\tikzmark{t2}*)
  (*\tikzmark{c2a}*)val findings =
    topics.par.map(topic =>
      agent[String](s"Research: $topic"))
  agent("Generate a report from the findings")(*\tikzmark{c2b}*)
\end{code}
\holeframe{c2a}{t2}{c2b}{c2b}{blue!55!black}{recursive expansion}

\noindent Each nested call is checked like the outer one, with its
own result type and access to the variables introduced by the code
around it. Recursive model calls are not a separate agent
protocol: they are ordinary code that can branch, loop, spawn
sub-agents, call skills, or route work across models.

The check catches structural failures before execution: a snippet
that uses a missing tool, passes arguments of the wrong shape, or
returns the wrong kind of result is rejected as a whole, and the
retry starts from an unchanged state (\Cref{sec:safety}).
The check also bounds the agent's
authority~\cite{dottycc,tacit,whatsinthebox}:
whether it can access certain files, network handles, 
and tools (\Cref{sec:capabilities}).

Our contributions are:
\begin{enumerate}
	\item A code-as-action model in which an LLM writes agent
	      actions as code that runs as part of its own runtime and
        is checked at the call site before
	      execution (\Cref{sec:model}).
	\item A safety analysis of the resulting guarantees:
	      pre-execution rejection of unavailable names and type
	      mismatches, no partial execution of rejected snippets, and
	      permissions and information-flow control
	      (\Cref{sec:safety}).
	\item A demonstration that nested \texttt{agent} calls and
	      ordinary code express common agent patterns, including
	      ReAct loops, skills, and multi-model planning,
	      as ordinary program control flow (\Cref{sec:arch}).
	\item A Scala~3 realization and an evaluation on a collection
	      of verifier test cases, BrowseComp-Plus~\cite{browsecompplus}, and $\tau^2$-bench~\cite{tau2bench},
	      including the retry behavior induced by compiler
	      diagnostics (\Cref{sec:realization}, \Cref{sec:eval}).
\end{enumerate}
\section{Related Work}
\label{sec:related}

Code-as-action approaches~\cite{codeact,codeexecution,smolagents}
let the model write code as its action space rather than emit a
single tool call. Recursive language models (RLM)~\cite{rlm},
introduced above, are the closest prior design to ours, and we
improve on it in two ways. First, RLM's REPL (a read-eval-print loop, the interactive
shell that retains state across inputs) runs generated code
without checking it first, so a snippet that misuses a binding or
returns the wrong shape can fail partway through and leave the
environment inconsistent, whereas \methodname{} typechecks against
\lstinline|T| in the live lexical scope before any of it runs.
Second, RLM hands the model a handle to the context but keeps
orchestration and control flow in the runtime, whereas in
\methodname{} the generated code writes that control flow itself, as
typed code over the \texttt{agent} primitive.

Other language-integrated LLM frameworks make model calls
first-class in a host language, but they focus on the model's input
and output rules.
LMQL~\cite{lmql} casts LLM inference as a query whose holes are
filled by constrained decoding, where declared constraints on the
type, length, or form of the result steer the sampler.
DSPy~\cite{dspy} describes an LLM call with a typed \emph{signature}
that the framework renders into a prompt and parses back into values.
In both, the declaration governs only a single call's input and
output. Composing several such calls into a larger workflow is left to
the developer, who wires them together by hand in fixed code,
such as an LMQL query or a DSPy pipeline.

\methodname{} differs on both counts.
The agent emits a \emph{program} rather than a constrained
string or a set of field values, and the host compiler
typechecks that program against \lstinline|T| in the call site's
lexical scope before it runs.
We neither constrain the sampler nor parse the output. Instead, the
compiler's error messages are fed back to drive retries until the
model produces a well-typed snippet.
Composition is then expressed by the generated code itself,
as control flow over the \texttt{agent} primitive.
And because the snippet is real code of the host language,
capture checking bounds the capabilities it may use, a guarantee
neither output-shaping framework provides.

The closest work in framing is ChatLSP~\cite{chatlsp}, which
likewise fills a typed hole with LLM-generated code from its expected
type and context. Its setting, though, is \emph{edit-time} code
completion that a human reviews, where context mainly reduces
hallucination. \methodname{} instead makes the hole a recursive
\emph{runtime} action, typechecked against the live lexical scope and
run in one process. That shift adds guarantees that completion does
not need: a dynamic dependency on the live context, capture-checked
authority over effects and data, and recursive use of the hole as the
unit of dynamic control flow rather than a one-shot completion.

\section{\methodname{}: Typed Holes as Agents}
\label{sec:model}

\subsection{The Agent Call}
\label{sec:hole}

\methodname{} treats an agent request as a placeholder in code:
the surrounding program needs a value whose type is fixed
\emph{statically}, and the model writes the code that should produce
it. Programming tools often call such a placeholder a \emph{typed
hole}~\cite{hazel}. We reuse the idea for model-written actions at
runtime:

\begin{code}
def agent[T](task: String): T
\end{code}

\noindent The type parameter \lstinline|T| is the expected result
type, and the value parameter \lstinline|task| is a natural-language
prompt describing what should go there. In practice, \lstinline|T|
rarely needs to be written out. Scala's type
inference picks it up from the surrounding context 
, so callers usually write \lstinline|agent(...)| and let the 
compiler fill in the type. 
At runtime, the LLM receives the prompt together with the expected
type and the enclosing source at the call site, and returns a
string of Scala source intended to produce a \lstinline|T|. The
compiler checks that source statically against \lstinline|T|,
as if it had been written at the call site. If the check succeeds,
the snippet runs and the call evaluates to a value of type \lstinline|T|.
If it fails, the agent receives the diagnostics as feedback
and can try again.

The static type itself does not constitute our entire contribution,
since any typed language provides one. What matters is \emph{when} 
and \emph{against what} it is enforced.
A compiler for a statically typed language normally checks only
source the developer wrote, ahead of time, and gives no way to run a
string against the contract of the surrounding code while the program
executes.
\methodname{} provides that guarantee for model-written code:
the snippet does not exist until runtime, yet it is checked against
\lstinline|T| and the live lexical scope at the call site under the
same static rules as hand-written code, before any of it runs.
The generated action thus inherits the full strength of static
checking from the host language (\Cref{sec:safety}),
rather than a weaker runtime approximation.

Concretely, the prompt sent to the LLM is assembled from a small
template: a system instruction telling the model to return a Scala
expression, the expected type \lstinline|T| rendered back to source,
the enclosing source with a placeholder at the \texttt{agent}
call's position, a
listing of the variables and parameters available at the call site
and their types, and the user's \lstinline|task| string.
The system instruction also carries
setup-specific guidance, for instance how to interact with the
user, how to request additional permissions or capabilities, and
how to organize a multi-step task into smaller \texttt{agent}
calls. The template is configurable per call site or per session.
Callers can swap the system instruction, change how available names are
summarized, or attach project-specific context for types the
model would not otherwise know.

\paragraph{What the model may write.} The generated code is typically 
a single expression or a block with multiple statements.
It may read parameters, read and update local
variables, use control flow
(\lstinline|if|-\lstinline|else|, \lstinline|while|, \lstinline|for|,
\lstinline|match|, \lstinline|try|-\lstinline|catch|), call any
function or method visible at the call site, including a
nested \lstinline|agent(...)|, or define its own local functions,
lambdas, or classes. 
The only requirements are the ones the compiler always enforces
for hand-written code: the final expression must have type
\lstinline|T|, every name it uses must be in scope at that point,
and the snippet must pass every other check the host compiler
applies.

\paragraph{Tools are functions.}
\label{sec:tools}
A \emph{tool} is simply a function in scope. The model invokes it by
writing a function call that the compiler type-checks, with no tool
registry, JSON schema, or protocol layer to maintain, and defining a
tool is just defining a function (see \Cref{sec:memory-tool}).
The idea extends to every interaction with the user and the environment,
so showing progress is a plain \lstinline|println(...)| and any I/O is
the corresponding standard-library call, with no separate agent layer
to mediate it.

\subsection{Examples}
\label{sec:examples}

The generated code is compiled as if the developer had typed it at
the exact point where the \texttt{agent} call appears. The snippet
can therefore use the same variables, functions, parameters, and
imports as hand-written code at that point. 

\begin{code}
> val xs = List(0, 1, 2, 4, 7, 9, 10)
> val r = agent[List[Int]](
|  "filter the prime numbers from xs")
// LLM produces:
//   def isPrime(n: Int): Boolean =
//     n > 1 &&
//       (2 until n).forall(n %
//   xs.filter(isPrime)
val r: List[Int] = List(2, 7)
\end{code}

\noindent The generated code uses \lstinline|xs| directly and defines
a local helper \lstinline|isPrime|. Because the snippet is compiled at
the call site, the name \lstinline|xs| refers to the list the
surrounding program defined, and the value is passed
to the snippet at runtime.
The expected type \lstinline|List[Int]| constrains the generation to
produce a list of integers, so the LLM cannot return a string, an integer, or a boolean.
The richer the result type, the tighter the contract:
\Cref{sec:more-examples} shows algebraic data types and
function types constraining the generated code further.

\subsection{Nested Agent Calls}
\label{sec:recursive}

Nested calls are the central mechanism of \methodname{}. The
top-level call \lstinline|agent[T](task)| asks the model for code
that solves the task, and that code may make smaller
\lstinline|agent[U](subtask)| calls. Each nested call has its own
expected type \lstinline|U| and its own task string, and is checked
and executed by the same \texttt{agent} mechanism.

Crucially, a nested call sees more than its parent did.
When the runtime reaches a nested \texttt{agent} call,
the LLM is asked to fill it within a richer context.
That context includes not only the names available at the
outer call site, but also every intermediate value, comment,
and control-flow structure the outer snippet has introduced up to that
point. Each sub-problem is therefore approached with more information.
The outer call has already narrowed the work down,
processed the relevant data, and recorded its reasoning in the program text.
Nested \texttt{agent} calls thus give an agent a natural way to break
a complex task into smaller ones, sequential or parallel, each reasoned
about with richer context and a more precise goal than the step before.
\Cref{sec:arch} shows that this is enough to express
common agent architectures.

Nested calls carry the usual termination caveat. An
LLM is free to emit a snippet that calls \texttt{agent} again, and
the new call may emit another, with no static bound on the depth.
A genuinely complex task and an accidental infinite recursion can
look the same from the outside. The runtime therefore tracks the
current depth of nested \texttt{agent} calls and exposes a
configurable cap. When the cap is hit, the offending call fails
with an exception. Callers who want a hard ceiling on cost
or latency set the cap themselves, and who prefer to trust the
LLM can leave it open and let the agent stop when it judges the
task complete.

\subsection{Handling Compilation Errors}
\label{sec:compile-errors}

Each \texttt{agent} call runs a self-correcting retry loop.
The generated code is sent through the compiler. If the check fails,
the diagnostics are appended to the original prompt and
the LLM is asked again, up to a configurable maximum number of
retries. If the agent still cannot produce an accepted snippet
within that budget, the call throws a special exception carrying 
the final compiler diagnostics. 
This is the appropriate failure when the prompt requests something
the surrounding program context cannot express,
for instance asking for a network call when no I/O capability is in
scope, or asking for a return shape the type system rules out.

The outer program can catch this exception like any other:

\begin{code}
try
  val x: Int = agent("...")
  ...
catch
  case e: EvalCompileException =>
    fallback()
\end{code}

\noindent The trade-off is that a \lstinline|try| block placed
around an outer \texttt{agent} call also catches compile failures
from any nested \texttt{agent} call inside its snippet, even when
those failures are unrelated to the outer call's intent.

\methodname{} also provides \lstinline|agentSafe[T]|, which, rather
than throwing on failure, returns its outcome as a value of type
\lstinline|EvalResult[T]| holding either the result value of type \lstinline|T|
or the final diagnostics. A caller can therefore handle a failed generation
locally instead of catching an exception that a nested call might
throw (the full signature of \lstinline|agentSafe| is in
\Cref{sec:realization}).

\section{Safety}
\label{sec:safety}

Each \texttt{agent} call is compiled by the host compiler in the
original lexical context, so a generated snippet is held to exactly
the rules the compiler applies to code written by hand at that point.
The snippet runs only if it resolves every name in scope, typechecks against
the expected \lstinline|T|, and passes every other check the compiler
enforces. These checks range from exhaustiveness and nullability to, when capture
checking is enabled, effect and information-flow constraints.
No separate safety pass of our own is involved: the guarantee is the
host compiler's soundness, applied to model-written code. We first
fix the threat model, then walk through representative rejections,
ending with the constraints capture checking adds. The fully
adversarial setting is developed in \Cref{sec:capabilities}.

In the examples below, the generated snippet appears as a comment
and the compiler's diagnostic is what the runtime reports to the
LLM and caller. We set the retry budget to zero so the first failure
surfaces directly (\Cref{sec:compile-errors}).

\subsection{Threat Model}
\label{sec:threat}

We make the trust boundary explicit. The \emph{trusted} components
are the compiler (type checker, static analysis, and code
generation), the runtime that executes a type-checked snippet, and
the host program that issues the \texttt{agent} call and supplies its
lexical scope. The \emph{untrusted} components are the model that
fills the hole (treated as potentially byzantine), every snippet it
produces, and any external content (files, third-party APIs, web
pages, and tool outputs) that reaches the task string.

The threat we address here is \textbf{model error}: even an honest,
well-intentioned model is an unreliable programmer and may emit code
that is incorrect or oversteps its bounds, e.g., performing I/O or
reaching a resource the surrounding program could not reach. We want
every generated snippet to be as safe as code a developer could have
written by hand at that point, irrespective of the model's
competence, so that a plain mistake never becomes an action outside
the snippet's static contract.
These guarantees hold against any model, honest or not, and
\Cref{sec:capabilities} extends them to a fully adversarial
one.

\subsection{Static Guarantees}
\label{sec:guarantees}

\paragraph{Undefined names.}
The snippet may use only names the lexical scope already provides.
A reference to a binding the surrounding program lacks is caught before
the snippet runs:

\begin{code}
> val tax: Double = 0.08
> agent[Double]("apply tax to price")
// model produces: price * (1.0 + tax)
<!EvalCompileException:
  agent failed to compile:
  Not found: value price!>
\end{code}

\paragraph{Type mismatches.} A value of the wrong
type cannot flow into a function call or an algebraic
constructor, even if the surface text looks plausible:

\begin{code}
> case class Order(id: Int, total: Double)
> val first: String = "A001"
> agent[Order]("id from first, total 0.0")
// model produces: Order(first, 0.0)
<!EvalCompileException:
  agent failed to compile:
  Order(first, 0.0)
        ^^^^^
  Found:    String
  Required: Int!>
\end{code}

The same checks turn away other common shortcuts.
\Cref{sec:more-rejections} shows a \lstinline|null| literal
rejected under explicit nulls~\cite{scala3explicitnulls} and a
non-exhaustive pattern match over a sealed data type rejected by the
exhaustiveness checker.

\paragraph{Atomicity: nothing runs if anything fails.} The
critical property is that the snippet is accepted or rejected
\emph{as a whole}. A side-effecting statement earlier in the
snippet does not run when a later statement fails to typecheck.
Consider an agent asked to update a mutable balance:

\begin{code}
> var balance: Int = 100
> agent[Int](
|  "subtract 50 and return the new balance")
// model produces:
//   balance -= 50
//   s"remaining: $balance" // String, not Int
<!EvalCompileException:
  agent failed to compile:
  Found:    String
  Required: Int!>
> balance
val res: Int = 100
\end{code}

\noindent The assignment to \lstinline|balance| precedes the
ill-typed expression in source order, yet never executes: the snippet
is rejected as a whole, so the runtime never runs its first
statement. Approaches that detect ill-typed code only at runtime (a
Python \lstinline|exec| string, an unconstrained tool call) leak
partial effects through exactly this pattern: the assignment fires
before the bad statement raises, leaving state inconsistent.
The typed hole is all-or-nothing by construction.

\subsection{Capability Safety}
\label{sec:capabilities}

The standard type system of \Cref{sec:guarantees} governs an
action's \emph{shape} but says nothing about its \emph{authority}:
which effects the generated code may perform and which data it may
use. Turning on Scala~3's capture checking
adds that layer as an opt-in, without changing the primitive,
building on the capture-set notation of
\Cref{sec:cc-primer}. A \emph{capability}, in the
object-capability sense used throughout this
paper~\cite{dennisvanhorn,objectcapabilites}, is an ordinary
unforgeable program value that authorizes a specific effect or
resource (a file handle, a network socket, a logger): code can
perform the effect only while it holds a reference to the
corresponding value. This differs from the systems notion of
capabilities as process-level privilege bits. Here, granting and
propagation are lexical and type-tracked rather than ambient.
The lexical scope at the hole is therefore the agent's permission
set, and the same compiler enforces capability scoping and
information-flow constraints on both the generated code and the
rest of the program.

\paragraph{Extended threat model.} The threat model of
\Cref{sec:threat} assumes an honest but fallible model.
Tracked capabilities let us widen it to an \emph{adversarial}
model that deliberately emits harmful code, the setting of
\emph{prompt injection}. An injection is \emph{direct} when it
rides in on a hostile user prompt and \emph{indirect} when it
arrives in content the agent reads at runtime (a tool result, a
file, or a fetched web page) that flows into the task string. We
do not try to stop the model from being \emph{influenced} by such
content, which is unavoidable once untrusted text reaches the
prompt. We only bound what that influence can do. The snippet is still
recompiled in the hole's lexical scope, so a subverted model can
invoke only the effects and reach only the data that scope already
grants, exactly as an honest model can.

\paragraph{Tool use as permission.} Because the scope is the
permission set, tool use is governed by the same mechanism. Under
capture checking an effectful tool (\Cref{sec:tools})
carries the capability it needs in its type, so the model can call
it only when the hole's scope binds that capability. Granting a
capability for one phase and withholding it in the next yields
least-authority, per-phase permissions: a snippet generated where
a read-only file handle is in scope can read, while one generated
without a network capability cannot send what it read, whatever a
poisoned instruction demands. The two results below make this
precise: confinement by scope bounds \emph{which} effects a
snippet may perform, and information-flow control bounds
\emph{which} data it may carry out.

\paragraph{Confinement by scope.} Capture checking confines a
capability to the region of code where it is lexically in scope. A
generated snippet can reach a capability only if the hole's context
already binds it, and even then it cannot smuggle that capability
out: a result whose type does not admit the capability cannot carry
it past the scope. The snippet may therefore invoke an in-scope
capability while it runs, but it may not hand back a value that
retains the capability for later use. An \lstinline|io| capability
makes the contrast concrete:

\begin{code}
> trait IO extends caps.SharedCapability
> def withIO[T](op: IO^ => T): T =
|  op(new IO {})
> def readFile(
|  io: IO, path: String): String = ...

// direct use: the result is a plain String
> withIO[String] { io =>
|  agent("read /etc/hosts using io")
| }
// model produces: readFile(io, "/etc/hosts")
val res0: String = ...

// storing io for later leaks it past the block
> withIO[String => String] { io =>
|  agent("return a file reader using io")
| }
// model produces:
//   (p: String) => readFile(io, p)
<!EvalCompileException:
Type Mismatch Error:
Capability io outlives its scope: it leaks into
outer capture set s1 owned by value res2.
The leakage occurred when trying to match the
following types:
Found:    String ->{io} String
Required: String ->s1 String!>
\end{code}

\noindent The first call uses \lstinline|io| directly and returns a
plain \lstinline|String|, which carries no capability, so it is
accepted. The second call asks for a \emph{function} that reads a
file. The generated lambda has type \lstinline|String ->{io} String|
because it captures the capability. The capability \lstinline|io| is
created fresh inside \lstinline|withIO| and scoped to that block, so it
cannot appear in the capture set of \lstinline|withIO|'s result,
which lives outside the block. Capture checking therefore
reports that \lstinline|io| \emph{outlives its scope}: the lambda would
carry \lstinline|io| out past the \lstinline|withIO| block that introduced
it, and the compiler rejects the leak before the body runs.

\paragraph{Information flow control: classified data.}
Capability scoping also rules out information flows that a
pure access-control language cannot describe. Consider a skill
that walks through several analysis steps over a legal contract.
The user holds a sensitive document and wants to run the skill,
but the \texttt{agent} in the program is powered by a hosted
online model, so sending the contract text to that model would
leak it.

The TACIT harness~\cite{tacit} addresses this with a typed
container. Sensitive content arrives wrapped:

\begin{code}
class Classified[T]:
  def map[U](f: T -> U): Classified[U]
\end{code}

\noindent The only way to touch the content is through
\lstinline|map|, which accepts a \emph{pure} function
(\lstinline|T -> U|, capture set empty). A pure function holds no
capabilities, so its body cannot read a file, open a socket,
print, or feed data back to the hosted model. The result of
\lstinline|map| is again \lstinline|Classified[U]|, so the wrapping is
preserved across the pipeline. The hosted agent can plan the
analysis, but the content never leaves the wrapper.

The limitation is that the function passed to \lstinline|map| is
written once, at code-generation time, so it cannot adapt to what
is inside the wrapper. The \texttt{agent} primitive lifts this
restriction: inside the function passed to \lstinline|map|, a nested
call to \lstinline|local.agent[U]| dispatches to a \emph{trusted}
local model. The nested call runs in the pure scope of
\lstinline|map|, so capture checking still rejects any effectful leak
of \lstinline|content|, but the code that processes the content is
now generated at runtime, with the content in view of the local
model alone:

\begin{code}
val doc: Classified[String] = docs.load(id)

val report: Classified[Report] =
  doc.map { content =>(*\tikzmark{patop}*)
    (*\tikzmark{pala}*)local.agent[Report](
      s"follow the skill steps on $content")(*\tikzmark{paend}*)
  }
\end{code}
\holeframe{pala}{patop}{paend}{paend}{blue!55!black}{pure action only}

\noindent The outer (hosted) agent generates the surrounding
program, including the lambda passed to \lstinline|map|. It sees the
\emph{source} of that lambda but not the value of
\lstinline|content|, which is bound only when \lstinline|map| fires at
runtime. At that point, \lstinline|content| reaches \lstinline|local|,
the trusted on-device model. The inner snippet is recompiled in
the same pure scope as the \lstinline|map| body, so capture checking
forbids it from invoking network IO, the file system, or the
hosted model's API. The outer agent plans without ever seeing the
content; the local agent acts on it with no way out. The net
effect is more flexibility without losing safety: a fixed pure
function commits the pipeline to one shape in advance, while a
nested typed-hole call lets the program adapt to the content at
runtime under the same capture-check argument.

We evaluate these guarantees empirically against an adversary in
\Cref{sec:agentdojo}, porting AgentDojo's prompt-injection
attacks to \methodname{}.

\subsection{Residual Escape Hatches}
\label{sec:hatches}
Two constructs slip past the type-level contract. \emph{Reflection}
lets code look up and use classes, fields, and methods by name at
runtime, reaching members and call paths the static checker never
sees. \emph{Raw process execution} launches an external process that
runs outside the programming language altogether.
Both are \emph{ambient authority}: any code can reach them without
being granted a capability, so a snippet can use them while holding
none. The language feature \emph{safe mode}~\cite{scala3safemode}, a
compiler setting that forbids exactly these unsafe constructs in
source, closes both. We therefore recommend running agents under safe
mode. Without it, these authorities stay open and a snippet must be
treated as ordinary untrusted code. Resource exhaustion,
non-termination, and latency lie outside the type system entirely and
are bounded by runtime budgets (see the \nameref{sec:limitations}). %
\section{Modeling Agent Patterns with \methodname}
\label{sec:arch}

With \texttt{agent} as the only new primitive, common agent
patterns reduce to plain control flow over the typed-hole shape.

\subsection{Typed Skills and Self-Improvement}
\label{sec:skills}

A \emph{skill} is the reusable unit through which an agent
encodes domain expertise. Existing solutions sit at two extremes.
At one end, the dominant approach, exemplified by Anthropic's Agent
Skills~\cite{agentskills}, ships a skill as a text-based guide the
agent consults: it carries domain knowledge well but is
unenforceable, since nothing stops the model from skipping a step
or deviating from the procedure. At the other end,
Voyager~\cite{voyager} and tool-maker systems~\cite{latm} store a
skill as a fixed piece of code: reproducible, but committed to a
single program that cannot adapt to new situations.

Our \texttt{agent} primitive lets a skill sit between these
extremes. A skill is an ordinary typed function: its signature is
fixed and checked end to end by the compiler, while its body,
generated per call, may freely mix plain code with nested
\texttt{agent} calls. 
The following example sketches a skill for reviewing a code change, 
which can sit anywhere on the spectrum, from fully delegated to fully coded:

\begin{code}
// 1. fully delegated: one agent call
def reviewPR(diff: Diff): Review =
  agent("apply the code-review checklist")
// 2. mixed: code skeleton, agent inside
def reviewPR(diff: Diff): Review =
  val critical = diff.files.filter(_.risky)
  val notes = critical.map { f =>
    agent[Note](
      s"review $f against the checklist")
  }
  Review.fromNotes(notes)
// 3. fully coded: no agent call
def reviewPR(diff: Diff): Review =
  val notes = diff.files.map(check)
  Review.fromNotes(notes)
\end{code}

\paragraph{Self-improvement.} A skill library is just a set of
functions in scope, so revision falls out of name resolution. In a
long-running REPL session, an agent emits a new
function with the same name and signature. This definition shadows
the previous one, so later \texttt{agent} calls resolve to the
updated version. The effect is to treat what
cognitive-architecture accounts~\cite{coala} call \emph{procedural
memory} as code edited in place, the role that
Reflexion~\cite{reflexion} and Voyager-style
libraries instead assign to free-form text or stored scripts.

\subsection{ReAct Loop}
\label{sec:react}

A ReAct~\cite{DBLP:conf/iclr/YaoZYDSN023} loop interleaves
reasoning and acting. Any loop can be written as a tail-recursive
function (one whose last action is a call to itself), so a ReAct
loop is naturally a tail-recursive \texttt{agent} call: at each
round the model emits a snippet that calls in-scope tools
(\Cref{sec:tools}), processes their results, and ends by
calling \lstinline|agent[T](task)| again, until it can return a
\lstinline|T| directly.
Conceptually, the call unwinds turn by turn:

\begin{code}
// initial agent call:
agent[T](task)
(*\tikzmark{r1top}*)// 1st round snippet:
(*\tikzmark{r1a}*)val x = readFile(f)
val y = parse(x)
agent[T](task)      // tail-call(*\tikzmark{r1end}*)
// 2nd round snippet:
val x = readFile(f)
(*\tikzmark{r2top}*)val y = parse(x)
(*\tikzmark{r2a}*)val z = analyze(y)
agent[T](task)      // tail-call(*\tikzmark{r2end}*)
// ... until the snippet returns a T directly
\end{code}
\holeframe{r1a}{r1top}{r1end}{r1end}{blue!55!black}{generated}
\holeframe{r2a}{r2top}{r2end}{r2end}{blue!55!black}{generated}

\noindent Every recursive call has the same expected return type
\lstinline|T|, so each turn attacks the same problem with more
accumulated context, and the loop ends when the surrounding scope
already contains enough information to produce a \lstinline|T|
without another model call.
This shape is closely related to recursive language modeling
(RLM)~\cite{rlm}, which also lets an agent write code that
re-invokes the model. 

Further patterns, such as sub-agents with an isolated context,
parallel reasoning over a collection, and planning with task
assignment across models, follow the same control-flow recipe and
are deferred to \Cref{sec:more-patterns}. %
\section{Realization in Scala~3}
\label{sec:realization}

The \texttt{agent} primitive is built on a lower-level operation,
\lstinline|eval[T](source)|, that takes a string of Scala source code
and runs it as if it had been written at the point where the call
appears (its \emph{call site}). The central technical challenge of
this work is to support such a dynamic operation, running code that is
only known as a string while the program is already running, inside a
statically typed host language without giving up the guarantees that
static typing provides.

\subsection{Why \texttt{eval} is Hard in a Static Language}
\label{sec:eval-hard}

Dynamic languages offer this operation for free. Python's
\texttt{eval(s, globals, locals)}~\cite{pythonexec} and JavaScript's
\texttt{eval(s)}~\cite{mdneval} take a string, turn it into code at runtime,
run it in the current scope or in a dictionary of variables passed
explicitly, and return whatever value results. The string is never
checked in advance: it runs in the same untyped setting as the rest of
the program, and any error surfaces only when the offending line
executes.

Reproducing this convenience in a statically typed host raises three
obstacles. First, a static compiler normally turns source into code
once, before the program runs, and does not compile arbitrary strings
afterward. To evaluate a string, the program must invoke the compiler
again, on itself, while it is already running. Second, the string must
be compiled as though it appeared at the call site, with access to
everything visible there: local variables, \emph{given} instances
(Scala's implicit values, which the compiler supplies from the
surrounding context), and capabilities (values that grant permission to
perform an effect such as I/O). In a dynamic language these bindings
sit in a runtime dictionary the interpreter can look up; in a static
language they live only in the compiler's internal representation of
the program, the syntax tree, and are gone by the time the program
runs. Third, this second, inner compilation must apply exactly the
same typing rules as the original one, so that the model-written code
is held to the same contract as the code around it. In particular,
capture checking (Scala's mechanism for tracking which effects and
resources a piece of code may use) must see the snippet in its
original context. Dynamic interpreters sidestep all three obstacles by
giving up static typing in the first place. Our goal is the opposite:
to keep the safety guarantees of a static host.

\subsection{The \texttt{eval} Primitive}
\label{sec:eval-primitive}

We add \texttt{eval} as a new built-in operation in the Scala~3
compiler. It takes two forms:

\begin{code}
def eval[T](source: String): T
def eval[T](
    code: String,
    bindings: Array[Binding],
    expectedType: String,
    enclosingSource: String
): T
\end{code}

\noindent The programmer writes only the short form,
\lstinline|eval[T](source)|: it takes a string of source code,
type-checks it against the expected type \lstinline|T| using
everything in scope at the call site, and runs it. The compiler then
expands this short form into the long one, filling in three pieces of
context automatically: \lstinline|bindings|, the in-scope variables
paired with their runtime values; \lstinline|expectedType|, the type
\lstinline|T| written back out as text; and \lstinline|enclosingSource|,
the text of the surrounding code with the call's location marked by a
placeholder. If the code string fails to type-check, \lstinline|eval| throws
an \lstinline|EvalCompileException| carrying the compiler's error
messages. A variant, \lstinline|evalSafe[T]|, instead returns the
outcome as a value of type \lstinline|EvalResult[T]|, an algebraic data
type with two cases, \lstinline|Success(value)| carrying the generated
value of type \lstinline|T| and \lstinline|Failure(diag)| carrying the
final compiler diagnostics, so the caller can treat a failed
compilation as ordinary data rather than catch an exception. The
user-facing \lstinline|agentSafe[T]| wraps \lstinline|evalSafe| the
same way \lstinline|agent| wraps \lstinline|eval|.

\subsection{Building \texttt{agent} on \texttt{eval}}
\label{sec:agent-on-eval}

The \lstinline|agent[T](task)| primitive and its sibling
\lstinline|agentSafe[T]| are thin wrappers around \lstinline|eval|,
written in ordinary Scala. They send the \lstinline|task| prompt,
together with the captured context, to an LLM; pass the Scala source
the model returns to \lstinline|evalSafe[T]|; and, when a compilation
fails, feed the compiler's diagnostics back into the next prompt and
try again. A simplified \lstinline|agentSafe| reads:

\begin{code}
@evalSafeLike
def agentSafe[T](
    task: String,
    bindings: Array[Binding] = Array.empty,
    expectedType: String = "",
    enclosingSource: String = "",
    maxAttempts: Int = 3): EvalResult[T] =
  @tailrec def loop(
      n: Int,
      errs: List[String]): EvalResult[T] =
    val prompt = buildPrompt(...)
    val code = llm.complete(prompt)
    val r = evalSafe[T](
      code, bindings,
      expectedType, enclosingSource)
    if r.isSuccess || n >= maxAttempts then r
    else loop(n + 1, r.error.errors.toList)
  loop(1, Nil)
\end{code}

\noindent The \lstinline|@evalSafeLike| annotation is what turns this
function into a hole. It instructs the compiler to fill the three
context parameters (\lstinline|bindings|, \lstinline|expectedType|,
\lstinline|enclosingSource|) at every call site, exactly as it does for
a direct \lstinline|evalSafe| call. Everything else is plain Scala: a
retry loop around \lstinline|evalSafe[T]| in which the LLM supplies the
candidate code and the compiler's diagnostics provide the feedback.

\subsection{How \texttt{eval} Works}
\label{sec:how-eval-works}

At a high level, code containing \lstinline|eval| (or an
\lstinline|evalLike| wrapper such as \lstinline|agent|) is transformed
and run in four steps:

\begin{itemize}
\item \textbf{Rewrite.} The compiler expands the \lstinline|eval|
  call into the long form, filling in the context parameters. This is
  done by a new compiler phase that runs after type checking, so it has
  access to the full typed syntax tree and can extract the necessary
  information from it.
\item \textbf{Splice.} At runtime, the splicer parses the source
  string and drops it into the placeholder in that
  enclosing-statement text, producing a complete top-level statement
  that looks exactly like code a developer could have written by hand
  at that spot.
\item \textbf{Recompile.} This spliced source is handed to a fresh
  run of the same compiler, configured with the same options
  (including the same effect and capability checks) as the original
  compilation. Type checking, capture checking, and error reporting
  are the ordinary ones the compiler applies to any program.
\item \textbf{Extract.} If the code compiles, the compiler produces a class file
  containing a method that evaluates the spliced code and returns its
  result. This class file is loaded into the running program.
\item \textbf{Evaluate.} The freshly compiled code is evaluated 
  in the program's original execution context
  (the same thread and class loader), yielding a value of type
  \lstinline|T| that \lstinline|eval| returns to the caller.
\end{itemize}

Reusing the unmodified compiler on the spliced source is what lets the
safety properties of \Cref{sec:safety} hold without any checker
of our own: the very guarantees the compiler provides for the
surrounding program apply to the generated code as well. All we have
done is arrange for the compiler to see the generated code embedded in
the right surrounding program.

\subsection{Portability}
\label{sec:portability}

Two ingredients of our prototype are specific to Scala: its
capture-checking system and the hook that lets a running program
invoke the compiler within its own process to splice and recompile.
Neither is fundamental. The design carries over to any statically
typed host that tracks effects or capabilities, as long as it can
recompile code from within a running program and expose the live
context at a call site to the rewriting step. %
\section{Evaluation}
\label{sec:eval}

Our evaluation asks four questions: do the type system's guarantees
hold across host and generated code (\Cref{sec:basic-tests});
can the recursive \texttt{agent} design handle complex, tool-using
tasks (\Cref{sec:browsecomp}); does it support multi-turn
conversation with tools (\Cref{sec:tau2}); and do the capability
guarantees of \Cref{sec:capabilities} hold against an adversary that
plants prompt-injection attacks (\Cref{sec:agentdojo})?
We realize \methodname{} as a Scala~3 library (\Cref{sec:realization}).
Full setup details, models, and resource budgets are described in \Cref{sec:setup}.

\subsection{Type-System Protection: A Collection of Test Cases}
\label{sec:basic-tests}

We isolate the type system as a verifier, independent of any model,
with roughly \(400\) test cases emulating how host code and a
generated snippet combine. Each pairs host code with a snippet that is
either \emph{well-formed} (should compile and evaluate to the expected
value) or \emph{ill-formed} in a way agents commonly emit (should be
rejected before it runs).
All tests pass, confirming that the implementation behaves as intended.

\subsection{Complex Tool-Using Benchmark}
\label{sec:browsecomp}

\begin{table}[t]
  \centering
  \small
  \begin{tabular}{lrrr}
    \toprule
    \textbf{Agent model} & \textbf{Acc.\ (\%)} & \textbf{Recall (\%)} & \textbf{\#retry} \\
    \midrule
    \texttt{deepseek-flash}     & 27.1 & 34.5 & 0.7 \\
    \texttt{gemini-lite} & 26.2 & 27.9 & 0.4 \\
    \texttt{gpt-mini} & 9.2 & 16.2 & 0.5 \\
    \bottomrule
  \end{tabular}
  \caption{\methodname{} on BrowseComp-Plus, varying only the
  agent model that powers the \texttt{agent} hole; corpus, retriever
  (\(k{=}5\)), and \texttt{gpt-4.1} judge are fixed. \emph{Acc.} is
  judge-scored correctness and \emph{Recall} is retrieval recall against the
  evidence documents, an upper bound on accuracy under the fixed
  budget; \emph{\#retry} is the mean compile-rejected regenerations per query.
  }\label{tab:browsecomp}
\end{table}

We test whether the recursive \texttt{agent} design serves as a
general agent on complex, tool-using tasks by running \methodname{} on
BrowseComp-Plus~\cite{browsecompplus}, a benchmark of hard
information-seeking tasks over a fixed corpus. Each task is a single
\texttt{agent} call that searches, reads results, and either answers
or recurses. Our aim is not to top the leaderboard, where accuracy is
dominated by retriever and base-model strength (both orthogonal to our
contribution), but to show that the typed primitive is a drop-in
general agent. \Cref{tab:browsecomp} shows that
\texttt{deepseek-v4-flash} answers \(27.1\%\)
correctly while driving genuine multi-step research (\(5.9\) rounds,
\(15.5\) searches per query), \texttt{gemini-3.1-flash-lite}
matches it at \(26.2\%\), and low-effort
\texttt{gpt-5.4-mini} explores little and scores
accordingly~\cite{deepseekv4,gemini31,gpt54mini}.
The primitive executes whatever agentic behavior the model generates.

The last column is the distinguishing one. Every snippet is
typechecked against its hole's contract before it runs, and one that
fails never executes: the compiler rejected \(8.6\%\) of generations
and returned the diagnostics for regeneration
(\Cref{sec:compile-errors}). The loop converges quickly: \(0.7\)
retries per query on average, two-thirds of queries needing none, and a
\(91.4\%\) end-to-end compile-success rate. The discipline thus adds no
accuracy cost, is cheap to satisfy, and guarantees that no ill-typed or
out-of-scope action reaches the corpus.

\subsection{Multi-Turn Conversation Benchmark}
\label{sec:tau2}

\begin{table*}[t]
  \centering
  \small
  \begin{tabular}{llrrrrr}
    \toprule
    & & \multicolumn{5}{c}{\textbf{Solved (\%, full reward)}} \\
    \cmidrule(lr){3-7}
    \textbf{Agent model} & \textbf{System} & \textbf{Retail} & \textbf{Airline} & \textbf{Telecom} & \textbf{Tel.-WF} & \textbf{Overall} \\
    \midrule
    \multirow{2}{*}{\texttt{deepseek-v4-flash}}
    & Tool Calling  & 80.9 & 75.5 & 100.0 & 95.4 & 90.0 \\
    & \methodname{} & 58.8 & 70.0 & 88.6 & 83.3 & 76.0 \\
    \midrule
    \multirow{2}{*}{\texttt{gemini-3.1-flash-lite}}
    & Tool Calling  & 52.6 & 56.0 & 15.8 & 21.1 & 33.2 \\
    & \methodname{} & 43.9 & 38.0 & 14.9 & 24.6 & 29.1 \\ %
    \bottomrule
  \end{tabular}
  \caption{\methodname{} on $\tau^2$-bench: percentage of tasks solved
    per service domain. A conversation is a sequence of
    \texttt{agent} calls sharing one REPL session.
    The user simulator is fixed at \texttt{gpt-5.4} across agent models so the
    simulated customer is a constant. Agent models run at temperature \(0\) with
    reasoning disabled. \emph{Tool Calling} is $\tau^2$-bench's reference
    agent, which invokes the same domain tools through the
    native function-calling API rather than generated code.
    \emph{Overall} is solved-rate pooled over all \(392\) tasks.
    }\label{tab:tau2}
\end{table*}

We then test multi-turn conversation with tool use, where state is
carried across turns and the agent interleaves tool calls with
replies. On $\tau^2$-bench~\cite{tau2bench}, a benchmark of tool-using
conversations across customer-service domains in which the agent and a
simulated user act on a shared environment, we realize each
conversation as a sequence of \texttt{agent} calls that share one REPL
session. Each user message becomes an \texttt{agent} call evaluated in
that session, so everything prior turns introduced (questions,
replies, tool calls, printed output) stays in scope. Conversational
context is thus carried by the REPL itself, with no dedicated memory
mechanism.

Across the four domains (\(392\) tasks), \texttt{deepseek-v4-flash}
solves \(76.0\%\) of tasks outright, ranging from \(58.8\%\) on
\texttt{retail} to \(88.6\%\) on \texttt{telecom}
(\Cref{tab:tau2}). These are genuine interactions, averaging
\(5.7\) user turns and \(26.7\) tool calls per task on \texttt{retail},
and the result is on par with the reference tool-calling agent.
As in the single-turn study, every snippet is typechecked against its
hole's contract before it runs, and the retry loop absorbs the
failures: \texttt{retail} averages \(7.1\) regenerations per task and
recovers to a \(77.6\%\) end-to-end compile-success rate, so no
ill-typed or out-of-scope action ever reaches the shared environment.

Conversational code is, however, more error-prone than single-turn
research code, because it must combine parsed tool results, prior-turn
state, and policy-conditioned actions: \(22.4\%\) of
\texttt{deepseek-v4-flash}'s \texttt{retail} generations are rejected,
against \(8.6\%\) on BrowseComp-Plus. How often the verifier fires also
depends strongly on the model's coding ability. On the weaker
\texttt{gemini-3.1-flash-lite}, the overall solve rate is low
(\(29.1\%\)) yet stays close to the reference agent, matching it on
\texttt{telecom} (\(14.9\%\) versus \(15.8\%\)) and exceeding it on
\texttt{telecom-workflow} (\(24.6\%\) versus \(21.1\%\)): the gap is the
model's, not the primitive's. The small model simply struggles to write
correct code, with rejections rising from \(3.2\%\) on \texttt{retail}
to \(89\%\) on \texttt{telecom}.

We expect these results to improve: prompt optimization or fine-tuning
that teaches a model to write typed code as an agent, rather than emit
isolated tool calls, should raise both the solve rate and first-try
compile success.

\subsection{Capability Safety Under Prompt Injection}
\label{sec:agentdojo}

\begin{table*}[t]
  \centering
  \small
  \setlength{\tabcolsep}{4pt}
  \begin{tabular}{llcccccccc}
    \toprule
    \multirow{2}{*}{\textbf{Model}} & \multirow{2}{*}{\textbf{System}} & \multicolumn{2}{c}{\textbf{Banking}} & \multicolumn{2}{c}{\textbf{Workspace}} & \multicolumn{2}{c}{\textbf{Slack}} & \multicolumn{2}{c}{\textbf{Travel}} \\
    \cmidrule(lr){3-4} \cmidrule(lr){5-6} \cmidrule(lr){7-8} \cmidrule(lr){9-10}
    & & Utility & Attack & Utility & Attack & Utility & Attack & Utility & Attack \\
    \midrule
    \multirow{3}{*}{gemini-2.5-pro}
    & CaMeL  & 52.8\% & 0/144 & 53.8\% & 0/560 & 48.6\% & 0/105 & \phantom{0}1.4\% & 0/140 \\
    & TACIT & 56.94\% & 0/144 & 50.54\% & 0/560 & 40.00\% & 0/105 & 64.29\% & 0/140 \\
    & \methodname{} & 63.19\% & 0/144 & 56.25\% & 4/560 & 42.86\% & 2/105 & 63.57\% & 1/140 \\
    \midrule
    \multirow{3}{*}{o4-mini-high}
    & CaMeL  & 62.5\% & 1/144 & 81.4\% & 0/560 & 68.6\% & 0/105 & 74.3\% & 0/140 \\
    & TACIT & 59.03\% & 0/144 & 52.86\% & 0/560 & 47.62\% & 0/105 & 68.57\% & 1/140 \\
    & \methodname{} & 65.97\% & 0/144 & 55.54\% & 0/560 & 49.52\% & 1/105 & 71.43\% & 0/140 \\
    \bottomrule
  \end{tabular}
  \caption{\methodname{} on the four stock AgentDojo domains.
  Utility is the fraction of user tasks completed successfully;
  Attack is the number of injection trials in which the attacker's
  goal was achieved over total trials. CaMeL numbers are taken
  from~\cite{camel}; TACIT numbers are taken from~\cite{tacit};
  \methodname{} numbers are from our runs.}\label{tab:agentdojo}
\end{table*}

This study stresses the capability layer of
\Cref{sec:capabilities} against an adversary. We extend
the TACIT benchmark~\cite{tacit}, which evaluates capability
tracking on agent code, with tasks drawn from
AgentDojo~\cite{DBLP:conf/nips/DebenedettiZBB024}, a dynamic
environment that plants prompt-injection attacks in the tool
outputs an agent consumes. We port the AgentDojo task suites to
\methodname{}, giving each agent only the capabilities its task
requires through scoped closures (\Cref{sec:tools}), and
run the AgentDojo attack suite against the ported agents.
TACIT and its data are open-source under the Apache License~2.0, and
the AgentDojo task and attack suites we port are released under the MIT
license.
\section{Discussion and Future Work}
\label{sec:discussion}

\paragraph{A foundation, not a replacement.} \methodname{} is not
meant to replace existing agent architectures but to give them a
more flexible and safer foundation: a single typed primitive for an
agent's behavior, its interaction with data, and its multi-step
reasoning, all checked statically.
The patterns of \Cref{sec:arch} are available as ordinary control
flow over \texttt{agent}, and a developer is free to keep any specialized
machinery an existing harness handles well. A conventional ReAct loop~\cite{DBLP:conf/iclr/YaoZYDSN023},
for instance, may manage the history of a long conversation
more efficiently than threading it through nested holes,
while it can still use \texttt{agent} calls for the dynamic behaviors
where type or capability safety matters.

\paragraph{Refinement-typed holes.} A natural next
step is to let the expected type \lstinline|T| carry a refinement
predicate~\cite{liquidtypes,scalarefinement}, so the contract
constrains not just the \emph{shape} of the result but also its
\emph{properties} (for example, an integer within a bound, a list of fixed
length, or relational invariants linking inputs to outputs).
First-class refinement types for Scala~\cite{scalarefinement} make
this concrete in our host language. Checking the refinement at the
hole would further open the door to verified decoding, steering
generation toward values that provably satisfy the predicate, with
the predicate discharged by a verifier such as Lean~\cite{lean4}
or Stainless~\cite{stainless}.

\section{Conclusion}
\label{sec:conclusion}

We have proposed \methodname{}, a single primitive \lstinline|agent[T](task)|,
that treats an agent's action as a typed hole in the host program.
At runtime the LLM fills the hole with code, compiled against
the expected type \lstinline|T| in the original lexical context,
fusing program execution and model reasoning into one process.
Recursion and composition over this primitive suffice to express
common agent patterns, including tools, typed skills, ReAct loops,
and multi-model planning, as ordinary control flow.
Safety follows from the host language itself: the type system
enforces scope and result-shape constraints, and a rejected snippet
never runs.
Across a collection of verifier test cases, BrowseComp-Plus, and
$\tau^2$-bench, the primitive serves as a drop-in agent while the
compiler blocks every ill-typed or out-of-scope action before it
runs, with diagnostics driving retries. 
\section*{Limitations}
\label{sec:limitations}

\paragraph{Well-typed is not correct.} The compiler checks that a
generated snippet has the expected static type and respects the
capabilities in scope. It does not check that the snippet does the
right thing. A well-typed snippet can still implement the wrong
algorithm, call the wrong in-scope tool, or return a plausible but
incorrect value, and the retry loop of
\Cref{sec:compile-errors} is silent on all three: it
regenerates only when the compiler rejects the snippet, so a snippet
that compiles runs whether or not it is semantically correct. Our
guarantees concern the \emph{shape} and \emph{authority} of an
action, not its semantic correctness, and human review or test
oracles remain necessary for the latter. Narrowing this gap by
letting the expected type carry a refinement predicate, so the
contract constrains the result's properties and not just its shape,
is a direction we outline in \Cref{sec:discussion}.

\paragraph{Authority is only as tight as the granted scope.} The
capability guarantees of \Cref{sec:capabilities} bound a
generated snippet to the effects and data its lexical scope already
grants, and they hold even against a model that emits hostile code.
What they do not do is prevent the model from being \emph{influenced}
by injected content. They only bound what that influence can reach.
The protection is therefore exactly as tight as the scope the
developer hands each hole. A hole over-provisioned with capabilities
it does not need reopens the attack surface, and an attacker who
steers the model into misusing a capability the task
\emph{legitimately} grants is not blocked. Consistent with this, a
small number of injection trials still succeed in our adversarial
study (\Cref{sec:agentdojo}). Least-authority scoping is thus
a property the developer must supply: the type system enforces it but
does not infer it.

\paragraph{Dependence on the model's coding ability.} Because a
rejected snippet never runs, the safety guarantee holds regardless of
how capable the model is; \emph{progress} does not. An agent advances
only when the model can express its intended action as well-typed host
code, so a model that writes weak Scala pays a heavy retry tax or
fails to converge, while the guarantee merely keeps its broken
attempts from executing. The effect is sharp and model-dependent in
our experiments (\Cref{sec:tau2}): only \(3.2\%\) of
\texttt{gemini-3.1-flash-lite}'s \texttt{retail} snippets are
rejected, but \(89\%\) are on the harder \texttt{telecom} domain,
where its code generation degrades and most turns make no progress. We
conjecture that part of this gap is due to Scala being less
represented in pretraining data than Python, the usual language of
code-as-action agents, so the competence floor a typed host imposes
is higher.
Closing it calls for stronger or better-adapted code models rather
than changes to the primitive.

\paragraph{Latency and cost.} Each \texttt{agent} call pays for a
model completion and at least one compiler pass, and nested recursion
stacks both. Every retry adds a further completion and pass, so cost
scales with the rejection rate: modest where the generated code is
clean (\Cref{sec:eval}). A body that is
generated once and reused, such as a function-typed hole compiled a
single time and applied many times
(\Cref{sec:more-examples}), amortizes the compile, but cold
calls and tight per-element loops with ever-varying generated code
remain expensive. The approach targets agent loops in which a model
call already dominates latency, so the current implementation is a
poor fit for ultra-low-latency settings.

\paragraph{Host-language requirements.} The design assumes a
statically typed host with an effect or capability discipline and an
in-process recompile mechanism that can expose the live call-site
context (\Cref{sec:portability}). Dynamically typed hosts
obtain the splice but none of the safety, which is why the guarantees
do not transfer for free to the Python stacks most agent frameworks
build on. The base \methodname{} prototype needs only ordinary static
typing. The permission, effect, and information-flow controls of
\Cref{sec:capabilities} are an opt-in layer on top that
additionally requires Scala~3 capture checking, an experimental
language feature. In either mode, the safety story relies on a safe
mode that closes reflection and raw process execution. Without it,
those ambient authorities remain escape hatches and a snippet
must be treated as ordinary untrusted code (\Cref{sec:hatches}).
Because \methodname{} executes model-generated code as real, effectful
actions, any deployment that relaxes these defenses, by omitting safe
mode or over-provisioning capabilities, carries a corresponding risk of
harmful actions, whether from model error or prompt injection.

\paragraph{Termination and resource use.} Recursion depth, fuel, and
wall-clock or memory limits are not type-level properties. A genuinely
complex task and a runaway recursion are indistinguishable to the type
system, so the runtime falls back on configurable depth and retry caps
(\Cref{sec:recursive}) and external budgets, such as the
nesting-depth cap and the per-query wall-clock limit in our experiments
(\Cref{sec:setup}). These budgets bound cost and non-termination,
but they must be set by the user, and a cap set too low can abort a
legitimate long-horizon task.

\bibliography{references}

\clearpage

\appendix
\label{sec:appendix}
\crefalias{section}{appendix}
\crefalias{subsection}{appendix}
\crefalias{subsubsection}{appendix}

\section{Richer Result Types}
\label{sec:more-examples}

The richer the type, the tighter the contract. An algebraic data
type pins the shape of the result:

\begin{code}
> case class Person(
|  name: String, born: Int, field: String)
> val turing: Person =
|  agent[Person]("info about Alan Turing")
// LLM produces:
//   Person(
//     name = "Alan Turing", born = 1912,
//     field = "Computer science")
val turing: Person = Person(
  name = "Alan Turing", born = 1912,
  field = "Computer science")
\end{code}

\noindent The generated code is a constructor call. The
case-class arity and field types are part of \lstinline|Person|, so
the model cannot return a value with a missing field or a
wrong-typed one.

A function type asks the model for an implementation:

\begin{code}
> val toRoman: Int => String =
|  agent("convert 1..3999 to Roman numerals")
// LLM produces:
//   val pairs = List(
//     1000 -> "M", 900 -> "CM", 500 -> "D",
//     400 -> "CD", 100 -> "C", 90 -> "XC",
//     50 -> "L",  40 -> "XL", 10 -> "X",
//     9 -> "IX",  5 -> "V",   4 -> "IV",
//     1 -> "I")
//   (n: Int) =>
//     var x = n; val sb = StringBuilder()
//     for (v, s) <- pairs do
//       while x >= v do { sb ++= s; x -= v }
//     sb.toString
> (1 to 5).map(toRoman)
val res1: Vector[String] =
  Vector("I", "II", "III", "IV", "V")
\end{code}

\noindent Here, the generated code is a function of type
\lstinline|Int => String|, stored once and called many times. Because
the function value is generated and compiled once, it can be
reused with no further LLM calls or compilation overhead.

\section{Additional Static Rejections}
\label{sec:more-rejections}

Beyond undefined names and shape mismatches
(\Cref{sec:guarantees}), the standard type system rejects
further common shortcuts.

\paragraph{Null safety.} Under Scala~3's explicit
nulls~\cite{scala3explicitnulls}, a
\lstinline|null| literal is not assignable to a non-nullable type,
which catches a common shortcut the model might otherwise reach
for:

\begin{code}
> val name: String =
|  agent("default user name, else null")
// model produces: null
<!EvalCompileException:
  agent failed to compile:
  Found:    Null
  Required: String!>
\end{code}

\paragraph{Pattern exhaustiveness.} The compiler flags
non-exhaustive matches over sealed shapes, which catches a common
failure mode of generated dispatch logic:

\begin{code}
> enum Color { case Red, Green, Blue }
> def label(c: Color): String =
|  agent[String](s"name the color $c")
// model produces:
//   c match
//     case Color.Red   => "red"
//     case Color.Green => "green"
<!EvalCompileException:
  agent failed to compile:
  match may not be exhaustive.
  It would fail on pattern case: Color.Blue!>
\end{code}

\section{Defining a Tool: A Memory Tool}
\label{sec:memory-tool}

\Cref{sec:tools} states that a tool is just a function in
scope, so defining a tool is just defining a function. A simple
memory tool, for instance, is a mutable map with a few functions to
manipulate it:

\begin{code}
val memory: mutable.Map[String, String]
def setMemory(key: String, value: String): Unit
def getMemory(key: String): Option[String]
def searchMemory(
    query: String): List[(String, String)]
def deleteMemory(key: String): Unit
\end{code}

\noindent With these names in scope, the agent uses the tool by
writing ordinary calls to them. The compiler checks each call
against its signature, with no registry or schema to keep in sync.
Suppose an email tool \lstinline|sendEmail(to, subject, body)| is also
in scope. The agent first records a meeting, then, asked to email a
colleague about it, recalls the details and sends the message:

\begin{code}
> agent[Unit](
|  "remember the team sync is Friday at 3pm")
// LLM produces:
//   setMemory("team-sync", "Friday 3pm")

> agent[Unit](
|  "remind Alice about the team sync")
// LLM produces:
//   val when = searchMemory("team sync")
//     .headOption.map(_._2).getOrElse("TBD")
//   sendEmail("alice@corp.com", "Team sync",
//     s"Reminder: the team sync is $when")
\end{code}

\noindent The second task never mentions a time, so the snippet
first calls \lstinline|searchMemory| to look the meeting up, then feeds
the result into \lstinline|sendEmail|, composing two in-scope tools in
a single typed snippet. The compiler checks the composition end to
end: that \lstinline|searchMemory| yields a list of
\lstinline|(String, String)| pairs the snippet destructures correctly,
and that \lstinline|sendEmail| receives arguments of the right type,
with no schema mediating the two calls.

\section{Tracked Capabilities in Scala~3}
\label{sec:cc-primer}

A \emph{capability} is an ordinary value tied to an effect or
resource: a file handle, a network socket, a logger, or a mutable
store~\cite{DBLP:journals/pacmpl/BrachthauserSO20}. In the
object-capability model~\cite{dennisvanhorn,objectcapabilites},
code can perform an effect only if it holds a reference to the
corresponding capability. Capabilities are unforgeable and
propagate only by being passed as ordinary data. Scala~3's
capture checking~\cite{dottycc,DBLP:journals/toplas/BoruchGruszeckiOLLB23,whatsinthebox}
lifts this discipline into the type system by recording, in a
value's type, which capabilities the value can reach.

Capturing types have the form \lstinline|T^{x1, ..., xn}|, where
the \emph{capture set} \lstinline|{x1, ..., xn}| over-approximates
the capabilities a value of this type may use. A type with an
empty capture set, written simply \lstinline|T|, is \emph{pure}
and retains no capabilities. The shorthand \lstinline|T^| (for
\lstinline|T^{any}|) admits any capability.

Function types record the capabilities their bodies use. The
closure \lstinline|(s: String) => f.write(s)| has type
\lstinline|String ->{f} Unit| (shorthand for
\lstinline|(String -> Unit)^{f}|), which makes explicit that the
function uses the file capability \lstinline|f|. A function whose
declared type is \lstinline|T -> U| is pure: its body cannot
invoke any capability, and any attempt to do so is rejected
before the body runs. The takeaway for this paper is that a
function type is a static whitelist of what the body may invoke,
and the lexical scope at a program point plays the same role for
the code that appears there. If a capability \lstinline|c| is not
in the lexical environment at a hole, no code that fills the hole
can invoke \lstinline|c|.

\section{Additional Related Work}
\label{sec:related-appendix}

Beyond the frameworks compared in \Cref{sec:related}, 
we cover several further lines of work,
especially ones bearing on the safety guarantees of
\Cref{sec:safety} and the capability layer (\Cref{sec:capabilities}).

\paragraph{Agent sandboxing and isolation.} Container and VM
isolation, syscall filtering, and language-subset Python
interpreters~\cite{monty} confine a generated snippet but enforce
only at runtime: a half-executed script can leave the surrounding
state inconsistent. Permission-scoped coding
agents~\cite{claudecode,opencode} gate tool access at the agent
boundary but share this non-atomic failure mode. We enforce
pre-execution, at capability granularity, with atomic failure.

\paragraph{Schema and policy-based access control.} JSON-schema
function calling and tool protocols such as MCP~\cite{mcp} are
safe only over pre-registered tools, and composition is checked
tool by tool rather than end to end. Security analyses report tool
poisoning and cross-origin abuse at the protocol
boundary~\cite{DBLP:journals/corr/abs-2504-03767,systemssecfoundations}.
Runtime policy languages such as Cedar~\cite{cedar}, as deployed
in Amazon's Bedrock AgentCore Policy~\cite{bedrockpolicy}, pin
access to a fixed list of resources but cannot constrain
information flow inside a permitted operation.

\paragraph{Prompt-injection defenses.} Output filtering,
training-based hardening such as StruQ~\cite{DBLP:conf/uss/ChenPSW25},
and LLM-as-judge monitoring are probabilistic, and recent
adaptive attacks bypass them in
practice~\cite{DBLP:journals/corr/abs-2510-09023}, even as
design-pattern catalogs set out principled
mitigations~\cite{DBLP:journals/corr/abs-2506-08837}. Dual-LLM
mediation and capability-based dataflow
defenses~\cite{dualllm,camel,tacit} push toward static checking
by having agents write capability-annotated programs, but enforce
outside the host type system, at coarse granularity, and with
non-atomic failure. We make the agent action a typed hole checked
by the host compiler, pre-execution, at capability granularity,
with atomic failure.

\paragraph{Capability-safe and effect-typed languages.}
Object-capability languages~\cite{objectcapabilites}, effect
systems, Scala~3 capture
checking~\cite{dottycc,DBLP:journals/toplas/BoruchGruszeckiOLLB23},
and region or ownership systems all provide the underlying
discipline. We \emph{apply} that discipline to the agent action
boundary. The novelty is the application and the \lstinline|eval|
mechanism that preserves it.

\section{Additional Agent Patterns}
\label{sec:more-patterns}

Beyond the patterns in \Cref{sec:arch}, three more reduce
to plain control flow over the typed-hole shape.

\subsection{Chain of Reasoning}
\label{sec:chain}

Chain-of-thought reasoning~\cite{cot} is a sequence of \texttt{agent} calls
nested in each other's scope. The simplest form passes one
call's output directly into another's prompt:

\begin{code}
val answer: Answer = agent(agent(
  s"polish this prompt: $task"))
\end{code}

\noindent The inner call rewrites \lstinline|task| into a sharper
prompt, and the outer call consumes the rewrite. Each call has its
own expected type, so the compiler checks that the inner result
is a \lstinline|String| before it reaches the outer. The same shape
generalizes to longer chains: every call captures the bindings
introduced by earlier ones and operates on that richer context,
and the compiler keeps the chain coherent end to end. An output
that does not fit the next call's parameter type fails before
that call runs. The pattern covers prompt polishing,
classification routed to a specialist, and any case where one
model's output feeds another.

\subsection{Sub-Agent with Isolated Context}
\label{sec:subagent}

A call site sometimes wants to delegate to an agent without
sharing its full scope. In \methodname{}, this is just a top-level
function that wraps an \texttt{agent} call:

\begin{code}
def subAgent[T](prompt: String): T =
  agent[T](prompt)
\end{code}

\noindent Calling \lstinline|subAgent[T](prompt)| from anywhere in
the program runs an \texttt{agent} call whose lexical context is
the body of \lstinline|subAgent|, not the caller's. Inside that body,
the names in scope are \lstinline|prompt| together with the top-level
definitions (imports, package-level definitions) reachable from this
file. Local bindings, capabilities, and instance members visible to
the caller do not leak in. A function signature is the natural way to budget
context: pass through what the sub-agent should see, and nothing
else.

\subsection{Parallel Reasoning}
\label{sec:parallel}

Because each \texttt{agent} call is an ordinary Scala expression,
parallelism comes from ordinary Scala combinators. To summarize a
collection of documents independently, use \lstinline|par.map|:

\begin{code}
val summaries: List[String] =
  files.par.map { f =>
    val text = readFile(f)
    agent[String](s"summarize: $text")
  }.toList
\end{code}

\noindent No special branching primitive is needed. The same
shape covers fan-out and fan-in, map-reduce over a collection, and
tree search where each branch is an independent \texttt{agent}
call.

\subsection{Planning and Task Assignment}
\label{sec:planning}

Different \texttt{agent} calls can target different models: a
small fast model for routine sub-tasks, a larger one for
planning, a trusted local model for sensitive data, and an
untrusted public provider for the rest. Each is a configured
\texttt{agent} instance:

\begin{code}
val plan: List[Subtask] =
  large.agent(s"plan the steps to $task")
val parts = plan.map { sub =>
  small.agent[Result](s"handle $sub")
}
val report: Report = large.agent(
  s"synthesize $parts into a report")
\end{code}

\noindent The planner uses a powerful model to pick a strategy
and emit scaffolding code, including calls to \lstinline|small.agent|
for the sub-tasks. Cost and capability decisions are local to
each call, and the type system does not need to know which provider
is on the other end.

This shape generalizes the dual-LLM
design~\cite{dualllm,camel}, where a privileged planner that
never sees raw inputs orchestrates a quarantined doer that
handles the data. Any partition of calls across two or more
configured agents instantiates the same pattern, with the split
chosen per call rather than fixed by the framework. By itself
this is only a routing convention, and a misrouted call still leaks.
\Cref{sec:capabilities} turns the partition into an
enforced barrier, using capture checking and
\lstinline|Classified[T]| to prevent the planner agent from
observing content it is not allowed to see.

\section{Experimental Setup}
\label{sec:setup}

\paragraph{Test Suite.} We build \methodname{} on the
Scala~\texttt{3.9.0} compiler.
The roughly \(400\) test cases of \Cref{sec:basic-tests} 
are contributed as REPL tests in the Scala~3 compiler's own test suite,
so each case exercises the real pipeline (parse, typer, capture check).

\paragraph{BrowseComp-Plus.} A Python driver issues each of the
\(830\) queries to the Scala REPL as a single \texttt{agent[String]}
call and runs the tool calls the generated code makes against a
fixed retrieval index. The agent has exactly two tools in scope:
\texttt{search(query)}, returning the top \(k{=}5\) corpus hits with
snippets, and \texttt{getDocument(docid)}, returning one full
document. Retrieval is held fixed across all runs: an exact-search
index over the canonical BrowseComp-Plus
\texttt{Qwen3-Embedding-8B}~\cite{qwen3emb} vectors
(\(\approx\!100\)k documents), with queries embedded by the
same model. Answers are graded by a fixed
\texttt{gpt-4.1}~\cite{gpt41} judge,
so accuracy differences track only the agent model.
We compare \texttt{deepseek-v4-flash} (high reasoning
effort), \texttt{gemini-3.1-flash-lite} (high reasoning
effort), and \texttt{gpt-5.4-mini} (low reasoning
effort), each at the provider's default sampling temperature 
\cite{deepseekv4, gemini31, gpt54mini}.
Every query runs in its own REPL under a \(600\)s wall-clock budget,
with recursive \texttt{agent} nesting capped at depth \(128\).
We log each query's input, answer, and tool calls, and trace every
snippet the agent generates with the compiler feedback it receives.

\paragraph{$\tau^2$-bench.} We run all four customer-service domains
(\texttt{retail}, \texttt{airline}, \texttt{telecom}, and
\texttt{telecom-workflow}; \(392\) tasks), scored by $\tau^2$'s
programmatic reward with no LLM judge. 
A Python driver runs the conversation loop: 
it shuttles each turn between the $\tau^2$ simulated user and the Scala REPL,
evaluating one \texttt{agent} call per user turn, 
and forwards every tool call the generated code makes to the $\tau^2$ server.
The setup choice that matters most is how those tools reach the agent: 
each domain ships a \emph{fixed} typed facade, 
one Scala function per tool with fully typed signatures
and nested arguments rendered as \texttt{case class}es rather than raw JSON.
The user simulator is fixed at \texttt{gpt-5.4}~\cite{gpt54},
and the agent models, \texttt{deepseek-v4-flash} and
\texttt{gemini-3.1-flash-lite}, run at temperature~\(0\) with
reasoning disabled, matching the baseline tool-calling agent.
Per-task limits guard against non-termination: \(200\) environment
steps, \(40\) user turns, \(500\) backend tool calls (\(50\) per
turn), and a \(300\)s idle timeout. 
We run one trial per task, which is equivalent to a \texttt{num-trials=1}
setting in the original $\tau^2$ benchmark.
The baseline evaluations use the same models and settings with the official scripts.
We log the full conversation, tool calls, and reward of each task,
and trace all generated code with its compiler diagnostics.

\paragraph{Model size and budget.} All agent, judge, and
user-simulator models (\texttt{deepseek-v4-flash},
\texttt{gemini-3.1-flash-lite}, \texttt{gpt-5.4-mini}, \texttt{gpt-4.1},
and \texttt{gpt-5.4}) are hosted endpoints accessed through
their providers' APIs, and we deliberately use the smaller, lower-cost
tiers (\emph{flash}, \emph{lite}, \emph{mini}) as the agent. Where a
provider discloses architecture we report it: \texttt{deepseek-v4-flash}
is a \(284\)B-parameter Mixture-of-Experts that activates \(13\)B
parameters per token, and retrieval uses the open \(8\)B
\texttt{Qwen3-Embedding-8B}. Google and OpenAI do not publish parameter
counts for the Gemini and GPT models, so we cannot report those sizes. We perform no
training or fine-tuning, so all model use is inference. The only local
computation is the BrowseComp-Plus retrieval index and the
Scala compiler passes each \texttt{agent} call triggers. Per-run
resource budgets are capped as described above (a \(600\)s wall-clock
and depth-\(128\) limit per BrowseComp-Plus query, and the per-task
limits on $\tau^2$-bench).
Because all models are hosted endpoints, the experiments require no
local GPU.

\paragraph{Licenses and release.} We plan to release our \methodname{}
implementation, test suite, and evaluation harness under the Apache License~2.0. The
benchmarks we build on are open-source: BrowseComp-Plus and
$\tau^2$-bench are both under the MIT license.
Our use of these artifacts is limited to research evaluation,
consistent with their intended use, and we release our own artifacts
for research. Because the benchmarks we build on are MIT-licensed
rather than research-only, our derivatives carry no research-only
restriction.
 
\end{document}